\documentclass[letterpaper, 10 pt,journal, twoside]{IEEEtran}  % Comment this line out if you need a4paper

\IEEEoverridecommandlockouts                              % This command is only needed if 
\usepackage{graphics} % for pdf, bitmapped graphics files
\usepackage{epsfig} % for postscript graphics files
\usepackage{times} % assumes new font selection scheme installed
\usepackage{amsmath} % assumes amsmath package installed
\usepackage{amssymb}  % assumes amsmath package installed
\usepackage{amsfonts}
\usepackage[colorlinks=true]{hyperref}
\usepackage[table]{xcolor}

\usepackage{booktabs}
\usepackage{multirow}
\usepackage{graphicx}
\usepackage{subfigure}
\usepackage{cite}
\usepackage{float}
\usepackage{makecell}
\usepackage{algorithm}
\usepackage{colortbl}
\usepackage{hyperref}
\usepackage[noend]{algorithmic}
\usepackage{etoolbox}
\usepackage[flushleft]{threeparttable}
\usepackage{threeparttable}

\hypersetup{
    colorlinks=true,
    linkcolor=blue,
    citecolor=blue,
    urlcolor=blue,  
    anchorcolor=blue
}

\makeatletter
\patchcmd{\@makecaption}
  {\scshape}
  {}
  {}
  {}
\makeatletter
\patchcmd{\@makecaption}
  {\\}
  {.\ }
  {}
  {}
\makeatother
\def\tablename{Table}

\graphicspath{{img/}}

\usepackage{makecell} 

\begin{document}

\title{$\mathrm{S}^2$-VLA: Decoupling Semantic and Spatial Streams in Vision-Language-Action Models for Autonomous Driving}

\author{Jianguo Yu$^{\dagger}$, Rukang Wang$^{\dagger}$, Duanfeng Chu$^{*}$, Chen Wang, Renju Feng, Liping Lu
\thanks{This work is supported in part by the National Natural Science Foundation of China (52472438), the Key R\&D Program of Hubei Province (2024BAB033, 2025BAB081), the Science and Technology Project of the Department of Transport of Hubei Province (2025-69-3-5), the Wuhan Joint R\&D Project Under the National Transportation Power Construction Pilot Program (2025-2-7), and the Wuhan Municipal key R\&D Program (20250512030403).
(Jianguo Yu and Rukang Wang contributed equally to this work.) ~(\textit{Corresponding author: Duanfeng Chu.})}
\thanks{Jianguo Yu and Chen Wang is with the School of Mechanical and Electronic Engineering, Wuhan University of Technology, Wuhan,  Hubei, China. (e-mail: \href{mailto:yujg@whut.edu.cn}{\texttt{\small yujg@whut.edu.cn}}, \hspace{0.25em} \href{mailto:310912@whut.edu.cn}{\texttt{\small 310912@whut.edu.cn}})}
\thanks{Rukang Wang, Renju Feng and Duanfeng Chu are with the Intelligent Transportation Systems Research Center, Wuhan University of Technology, Wuhan,  Hubei, China. (e-mail: \href{mailto:wangrk@whut.edu.cn}{\texttt{\small wangrk@whut.edu.cn}}, \hspace{0.25em} \href{mailto:fengrenju@whut.edu.cn}{\texttt{\small fengrenju@whut.edu.cn}}, \hspace{0.25em} \href{mailto:chudf@whut.edu.cn}{\texttt{\small chudf@whut.edu.cn}})} 
  \thanks{Liping Lu is with the School of Computer Science and Artificial Intelligence, Wuhan University of Technology, Wuhan, Hubei, China. (e-mail: \href{mailto:luliping@whut.edu.cn}{\texttt{\small luliping@whut.edu.cn}})} 
  }

\maketitle
\pagestyle{empty} % no page number for the second and the later pages
\thispagestyle{empty} % no page number for the first page
%\markboth{IEEE Robotics and Automation Letters. Preprint Version. Accepted August, 2023}

%%%%%%%%%%%%%%%%%%%%%%%%%%%%%%%%%%%%%%%%%%%%%%%%%%%%%%%%%%%%%%%%%%%%%%%%%%%%%%%%

\begin{abstract}
Vision-Language Models (VLMs) have demonstrated remarkable potential for high-level reasoning in autonomous driving, yet they fundamentally struggle to generate precise, low-level control actions. This limitation is rooted in a semantic-physical gap caused by the inherent mismatch between discrete language tokens and continuous trajectory planning. While Vision-Language-Action (VLA) architectures attempt to bridge this gap by unifying perception and control into a single policy, this entanglement creates a new bottleneck. Standard VLAs experience a severe spatial representation collapse, which irreversibly degrades the fine-grained spatial and geometric priors essential for safe, boundary-aware navigation.
To address this limitation, we propose the $\mathrm{S}^2$-VLA, 
which explicitly decouples the semantic and spatial streams in Vision-Language-Action models.
The semantic stream leverages hierarchical bridging to extract multi-scale VLM features for robust intent reasoning. In parallel, an independent spatial stream bypasses the autoregressive language bottleneck, directly preserving uncompressed spatial features from the visual encoder. By integrating auxiliary perception supervision, this stream explicitly equips the model with rich spatial and geometric priors. Finally, a dual-stream planning adapter fuses high-level semantic intent with precise spatial constraints via cascaded attention mechanisms.
Evaluations on the NAVSIM closed-loop benchmark show that $\mathrm{S}^2$-VLA achieves a Predictive Driver Model Score (PDMS) of 87.1, establishing a new state-of-the-art for VLA models under a purely supervised fine-tuning (SFT) setting. By mitigating the spatial representation collapse of traditional VLMs, our framework significantly outperforms baselines, achieving the highest No Collision (NC) rate of 98.4 among all evaluated methods.
\end{abstract}


\section{Introduction}
\IEEEPARstart{E}{nd}-to-end (E2E) autonomous driving has fundamentally shifted the paradigm by directly mapping raw multi-modal sensor inputs to continuous vehicle control actions or waypoints \cite{du2026bevdrive}. By unifying perception, trajectory prediction, and motion planning within a single differentiable framework, E2E models learn shared latent representations that effectively mitigate the cascading errors inherent in traditional modular pipelines. This paradigm has evolved from early behavioral cloning approaches to unified multi-task frameworks such as UniAD \cite{hu2023uniad} and VAD \cite{jiang2023vad}, and has been further advanced by generative priors and Mixture-of-Experts routing, as exemplified by DiffusionDrive \cite{liao2025diffusiondrive} and ARTEMIS \cite{feng2025artemis}.

Despite these advances, conventional E2E models remain largely data-driven black boxes. Their limited semantic interpretability and lack of high-level reasoning hinder generalization in long-tail scenarios, weaken multi-agent interaction modeling, and restrict the integration of human instructions.

Vision–Language Models (VLMs) address these limitations by introducing semantically grounded reasoning. Trained on large-scale image–text pairs, VLMs align visual inputs with language in a shared representation space, enabling natural language understanding and Chain-of-Thought (CoT) reasoning conditioned on the scene. This cross-modal abstraction improves generalization in complex and long-tail scenarios \cite{wen2023dilu,mei2024continuously}. Recent works, such as DriveLM \cite{sima2024drivelm} and DriveGPT4 \cite{xu2024drivegpt4}, leverage CoT prompting to jointly generate driving decisions and interpretable rationales, achieving strong zero-shot performance. Follow-up work like  ReasonPlan \cite{liu2025reasonplan} further enhances transparency by producing step-by-step decision processes.

However, extending VLMs to low-level control exposes a fundamental \emph{semantic–physical gap} \cite{jiang2025survey}. VLM outputs are inherently discrete and autoregressive (e.g., “turn left” or “slow down”), whereas vehicle control requires continuous, high-frequency signals governed by trajectory planning. Moreover, tokenization and deep abstraction progressively degrade fine-grained spatial and geometric information critical for precise motion generation. As a result, VLMs typically operate as open-loop “advisors”, relying on downstream controllers to translate symbolic outputs into executable actions. This multi-stage design introduces latency and further loss of spatial fidelity, limiting control reliability.

To bridge the coherence gap between language and action, Vision-Language-Action (VLA) models unify perception, reasoning, and control within a single policy. Originating in robotics (e.g., RT-2 \cite{zitkovich2023rt} and VLA-Adapter \cite{wang2026vla-adapter}), VLAs encode visual observations, ego states, and language instructions into a shared latent space, where cross-modal attention infers intent and action heads generate continuous control signals \cite{renz2025simlingo, li2025recogdrive, zhou2025autovla}. Recent advances demonstrate the ability to ground free-form instructions into trajectories \cite{fu2025orion}, produce interpretable reasoning via CoT \cite{zhao2025cot}, and incorporate expressive planners such as diffusion-based action generators \cite{jiang2025diffvla}.

Nevertheless, existing VLA architectures introduce a critical structural limitation. Most approaches rely on supervised fine-tuning (SFT) to enforce action generation or align control outputs with highly abstract features from deep network layers. While effective for coarse behavior prediction, these strategies overlook the progressive loss of fine-grained spatial information. This issue is exacerbated by the prevalent single-stream design, which directly maps high-level semantic representations to low-level control, entangling reasoning with execution. Consequently, cross-modal alignment is performed on heavily compressed embeddings, neglecting the rich spatial cues preserved in earlier visual features and inducing a systematic semantic-over-geometry bias. Although such models can mimic expert behavior, their lack of explicit spatial and geometric priors limits spatial awareness, often resulting in unsafe or geometrically inconsistent trajectories.

\begin{figure}[!htbp]
    \centering
    \includegraphics[width=0.48\textwidth]{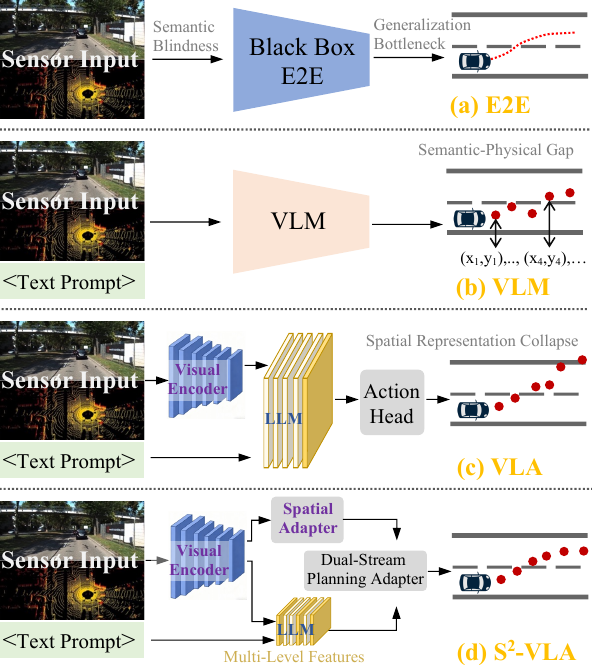}
    \caption{Overview of $\mathrm{S}^2$-VLA compared with existing paradigms. (a) Traditional E2E models suffer
from semantic blindness, lacking the interpretability and high-level cognitive
reasoning required for complex traffic scenarios.
 (b) VLMs produce discrete outputs unsuitable for continuous control. (c) Conventional VLAs suffer from spatial representation collapse, leading to loss of fine-grained spatial and geometric details. (d) Our proposed $\mathrm{S}^2$-VLA  enables safe, boundary-aware, and physically compliant trajectory generation.}
    \label{fig:limitation}
\end{figure}

As illustrated in Fig.~\ref{fig:limitation}, current paradigms face structural limitations. To address these challenges, 
we introduce $\mathrm{S}^2$-VLA, which decouples semantic and spatial processing streams in Vision-Language-Action models.
 Unlike conventional VLAs that entangle reasoning and control, $\mathrm{S}^2$-VLA explicitly decouples semantic reasoning from spatial perception, enabling safe, boundary-aware, and physically compliant trajectory generation.

The main contributions of this work are  as follows.

(1) We propose $\mathrm{S}^2$-VLA to overcome the \emph{semantic–physical gap} inherent in current models by fundamentally decoupling the semantic and spatial streams. By harvesting spatially dense features from shallow visual layers and applying auxiliary perception supervision, the  spatial stream explicitly grounds the model in rich geometric priors for accurate spatial reasoning.

(2) We design a Dual-Stream Planning Adapter that reconciles high-level semantic intent with low-level control. Through hierarchical cross-attention, it effectively fuses semantic guidance with strict spatial constraints, ensuring both logical consistency and geometric constraints.

(3) Extensive evaluations on the NAVSIM closed-loop benchmark demonstrate that $\mathrm{S}^2$-VLA achieves a Predictive Driving Model Score (PDMS) of 87.1, establishing a new state-of-the-art for VLA models under a purely supervised fine-tuning (SFT) setting. By mitigating the spatial representation collapse inherent to traditional VLMs, our framework significantly outperforms existing baselines, notably achieving the highest No Collision (NC) rate of 98.4.

\vspace{-1em}

\section{Method}
\begin{figure*}[!h]
    \centering
    \includegraphics[width=1\textwidth]{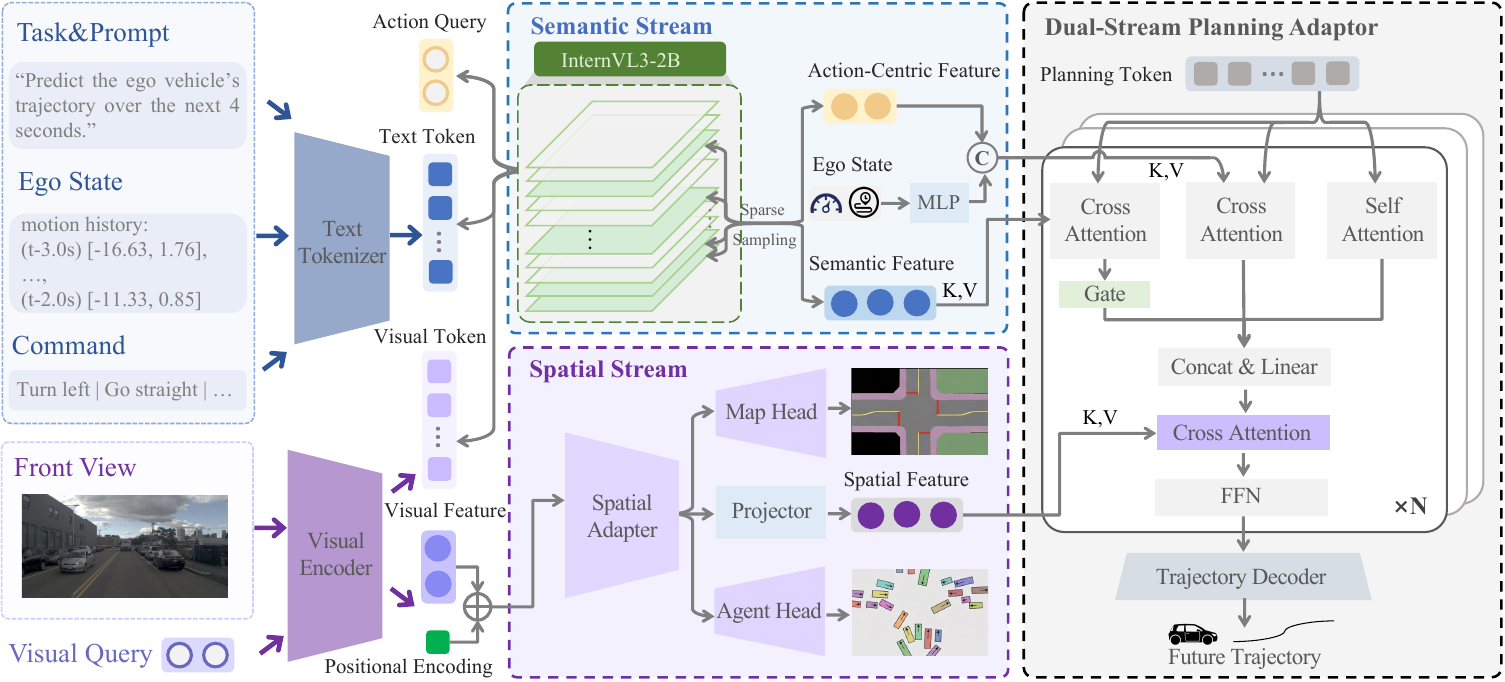}
    \vspace{-0.5cm}
    \caption{Architecture of the $\mathrm{S}^2$-VLA model.}
    \vspace{-0.3cm}
    \label{fig:model}
\end{figure*}

The overall architecture of the proposed  ($\mathrm{S}^2$-VLA) model is illustrated in Fig.~\ref{fig:model}. To resolve the semantic-geometric entanglement inherent in conventional VLA models, $\mathrm{S}^2$-VLA processes multimodal inputs, including a high-level navigation command $C_\text{nav}$, ego-vehicle motion history $T_\text{hist}$, a front-view image $I$, and learnable visual and action queries. Following initial tokenization and vision encoding, the architecture explicitly decouples representation learning into two  streams. 

The \textcolor[HTML]{70319f}{Semantic Stream} leverages InternVL3-2B~\cite{zhu2025internvl3} as its vision-language backbone. By jointly encoding text and visual tokens, it extracts high-level, multi-scale semantic features that encapsulate complex reasoning and driving intent. 

In parallel, the \textcolor[HTML]{476eba}{Spatial Stream} bypasses the autoregressive language bottleneck by directly extracting uncompressed visual features from the Visual Encoder. Furthermore, this stream preserves fine-grained spatial details and enforces explicit spatial and geometric priors via auxiliary map and agent prediction tasks. 

Finally, the \textcolor[HTML]{595959}{Dual-Stream Planning Adapter} aligns these decoupled representations via cascaded attention mechanisms, seamlessly fusing abstract semantic intent with precise spatial constraints to generate dynamically feasible trajectories.

\vspace{-1em}
\subsection{Multi-Scale Semantic Stream}
Current VLM-based driving systems primarily rely on the final-layer representations of their backbone networks. Due to heavy network compression, these highly abstract features suffer from semantic discretization and a distinct semantic-over-action bias. Consequently, they lack the fine-grained multimodal details necessary to ground high-level reasoning into precise, executable actions for trajectory planning. To overcome this limitation, we propose a Multi-Scale Semantic Stream designed to selectively extract and aggregate hierarchical features.

\textbf{Multimodal Input and Query Injection.} 
All inputs are tokenized into a unified sequence of text and visual tokens. At the input of the semantic stream, we inject $N_{\text{act}} = 64$ learnable action queries $Q_{\text{act}}$ to explicitly drive the network to aggregate execution-relevant features into action-centric representations. 
 The choice of $N_{\text{act}} = 64$ directly follows the configuration validated in VLA-Adapter~\cite{wang2026vla-adapter}. 
This number provides the optimal balance between representational capacity and computational efficiency.

\textbf{Backbone and Sparse Sampling.} 
We adopt InternVL3-2B~\cite{zhu2025internvl3} as the backbone, which combines the InternViT visual encoder with the Qwen2.5 language model to jointly process multimodal tokens and action queries. While full-layer feature fusion is computationally prohibitive, Transformer layers yield inherently complementary representations: shallow layers retain fine-grained details, while deeper layers capture high-level semantics~\cite{wang2026vla-adapter,chen2025multimodal,ciernik2026beyond}.

To balance inference efficiency with representation richness, we adopt a sparse multi-scale sampling strategy. Specifically, we extract hidden states exclusively from a carefully selected subset of layers, $L = \{3, 8, 13, 18, 23, 24\}$, forming the multi-scale  feature set
$ \big\{ (V_\text{sem}^{(l)}, V_\text{act}^{(l)}) \mid l \in L \big\}$,
where $V_\text{sem}^{(l)} \in \mathbb{R}^{(N_\text{text}+N_\text{img}) \times d_\text{vlm}}$ denotes  semantic  features at layer $l$, and $V_\text{act}^{(l)} \in \mathbb{R}^{N_\text{act} \times d_\text{vlm}}$ represents the corresponding action-centric features. $N_\text{text}$, $N_\text{img}$, and $N_\text{act}$ denote the sequence lengths of the text tokens, image tokens, and action queries, respectively, while $d_\text{vlm}$ is the hidden dimension of the backbone. 

\textbf{Output and State Integration.} 
The semantic stream outputs dense semantic features $V_{\text{sem}}$ and action-centric features $V_{\text{act}}$. To explicitly incorporate kinematic priors, the ego-vehicle’s historical states are encoded via a Multi-Layer Perceptron (MLP) into an ego-state embedding $E_{\text{ego}} \in \mathbb{R}^{1 \times d_{\text{vlm}}}$. At each layer $l$, this embedding is concatenated along the sequence dimension with the action-centric features $V_{\text{act}}^{(l)}$ to form a unified state memory $E_{\text{mem}}$. Together, this memory $E_{\text{mem}}$ and the dense semantic features $V_{\text{sem}}$ provide a structured, intent-aware representation that serves as the crucial semantic foundation for the downstream Dual-Stream Planning Adapter.

\vspace{-1em}
\subsection{Task-Driven Spatial Stream}
The autoregressive pipeline in VLMs inherently compresses visual representations, severely weakening explicit spatial awareness. To address this limitation, we design a dedicated Spatial Stream that  directly accesses visual features from the Vision Transformer (ViT) and injects explicit spatial and geometric priors through task-driven supervision.

\textbf{Visual Input and Feature Extraction.}
We adopt a dynamic resolution strategy~\cite{guo2024llava}, partitioning each high-resolution front-view image into $N_{patch}=9$ parallel-processed patches, consisting of eight local crops and one global thumbnail. To actively capture spatial information, we introduce $N_\text{vis}=64$ learnable visual queries $Q_\text{vis}$ per patch. 
These queries are concatenated with the standard class token $H_{\text{CLS}}$ and patch tokens $H_{\text{patch}}$, formulating the input sequence to the Vision Transformer (ViT) as $[Q_{\text{vis}}; H_{\text{CLS}}; H_{\text{patch}}]$.

\begin{figure}[!hbtp]
    \centering
    \includegraphics[width=1\linewidth]{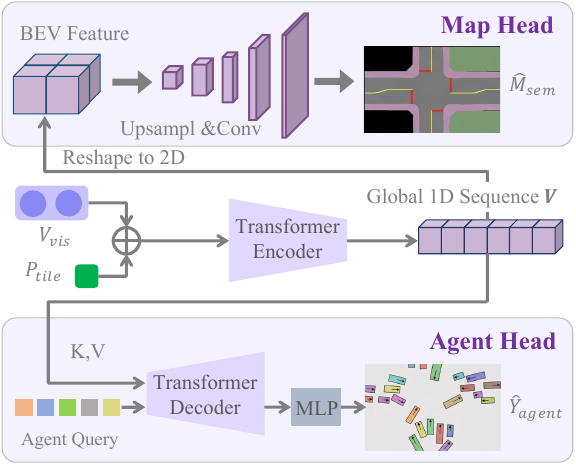}
    \vspace{-0.5cm}
    \caption{Architecture of the spatial stream.}
    \vspace{-0.3cm}
    \label{fig:spatial-stream}
\end{figure}

During the forward pass, $Q_{vis}$ aggregates fine-grained geometric and texture details via self-attention. After encoding, the visual queries are decoupled and reorganized into a spatially dense visual feature $V_{\text{vis}} \in \mathbb{R}^{B \times (N_{\text{patch}} \times N_{\text{vis}}) \times C}$, successfully preserving fine-grained geometric information. To establish global spatial coherence, we add tile-level positional embeddings $P_{tile}$ and apply a Transformer encoder, which is then explicitly grounded in spatial and geometric priors via two auxiliary perception heads, illustrated in Fig.~\ref{fig:spatial-stream}.

\textbf{Map Head.}
To explicitly establish a spatial geometric prior of the environment, the 1D token sequence $V$ is deterministically reshaped into a 2D spatial grid $V_{2D} \in \mathbb{R}^{d \times 24 \times 24}$. Subsequently, a progressive sequence of upsampling and convolutional layers decodes this intermediate grid to output a high-resolution BEV multi-class semantic map $\hat{M}_{\text{sem}} \in \mathbb{R}^{C \times 128 \times 256}$.
Because our architecture relies exclusively on a monocular front-view camera, the Map Head focuses on reconstructing a local, visible BEV semantic space immediately ahead of the ego-vehicle(specifically spanning $X \in [0, 32]$ m and $Y \in [-32, 32]$ m).
To mitigate the severe class imbalance between background pixels and safety-critical elements, this branch is supervised via a weighted cross-entropy loss:
\vspace{-0.1 em}
\begin{equation}
\mathcal{L}_{\text{map}} = -\sum_{c} \omega_c M_{gt}^{(c)} \log \hat{M}_{\text{sem}}^{(c)},
\end{equation}
where $M_{gt}^{(c)}$ is the ground-truth binary mask for class $c$, $\hat{M}_{\text{sem}}^{(c)}$ denotes the predicted semantic probability, and $\omega_c$ represents class-specific weights applied to emphasize critical topological features such as drivable areas and lane boundaries. Ultimately, this localized representation provides explicit physical references regarding road topology, serving as a crucial foundation for compliant trajectory planning.

\textbf{Agent Head.} 
To capture dynamic objects, we follow the DETR paradigm~\cite{carion2020detr} using a Transformer decoder equipped with $N_{\text{agent}}=30$ learnable queries. A MLP head decodes these queries into continuous agent predictions:
\begin{equation}
\hat{Y}_{\text{agent}} = \{\hat{b}_i, \hat{p}_i\}_{i=1}^{N_{\text{agent}}},
\end{equation}
where $\hat{b}_i$ parameterizes the predicted oriented 2D bounding box (position, size, and heading) and $\hat{p}_i$ indicates the predicted classification confidence. We employ the Hungarian algorithm to establish an optimal bipartite matching $\sigma$ between the predicted queries and the ground-truth obstacles. For the matched positive samples, we align the predictions with the ground-truth bounding box state $b_{gt}$ and the target class label $p_{gt}$. The total agent loss is a weighted sum of an $L_1$ spatial regression loss and a Binary Cross-Entropy (BCE) classification loss:
\begin{equation}
\mathcal{L}_{\text{agent}} = \lambda_{\text{reg}}\mathcal{L}_{L1}(\hat{b}_{\sigma}, b_{gt}) + \lambda_{\text{cls}}\mathcal{L}_{BCE}(\hat{p}_{\sigma}, p_{gt}),
\end{equation}
where $\lambda_{\text{reg}}$ and $\lambda_{\text{cls}}$ are hyperparameter coefficients used to balance the two loss terms. 

Through this multi-task supervision, the spatial stream is strictly regularized to encode metric layouts and spatial boundaries rather than arbitrary visual textures. Finally, the intermediate representation $V$ is linearly projected to form $V_{\text{spatial}}$, a dense, geometrically grounded feature. This representation serves as a rigorous spatial constraint for the downstream Dual-Stream Planning Adapter, guaranteeing that the generated trajectories are physically feasible and strictly aligned with environmental constraints.

\subsection{Dual-Stream Planning Adapter}
To bridge high-level semantic intent with low-level physical constraints, we introduce the Dual-Stream Planning Adapter, a cascaded multi-stage cross-attention decoder. This module progressively fuses representations from the semantic and spatial streams to produce trajectories that are both semantically aligned and physically feasible.

We initialize a set of learnable planning tokens $P^{(0)} \in \mathbb{R}^{M \times d}$, where $M=8$ represents the future trajectory waypoints and $d$ is the hidden dimension. The Adapter consists of $N$ stacked blocks. Each block refines the planning tokens through two sequential stages:

\subsubsection{Semantic and State Alignment}
Inspired by hierarchical bridging strategies~\cite{wang2026vla-adapter}, the cross-modal attention mechanism in this stage establishes coarse-grained navigation intent by leveraging the commonsense reasoning capabilities of the VLM. Specifically, the planning tokens are aligned with multi-scale semantic features $V_{\text{sem}}$ and ego-state memory $E_{\text{mem}}$ to capture both contextual intent and dynamic constraints. Given $P^{(l-1)}$ as queries, we perform parallel cross-attention and self-attention operations:
\vspace{-0.5 em}
\begin{equation}
\begin{aligned}
P_{\text{raw}} &= \mathrm{MHCA}_1(P^{(l-1)}, V_{\text{sem}}, V_{\text{sem}}), \\
P_{\text{ego}} &= \mathrm{MHCA}_2(P^{(l-1)}, E_{\text{mem}}, E_{\text{mem}}), \\
P_{\text{sa}}  &= \mathrm{MHSA}(P^{(l-1)}).
\end{aligned}
\end{equation}

A learnable gating parameter $g \in \mathbb{R}^{d}$ modulates the contribution of the semantic features. The fused intermediate representation is computed via linear projection and concatenation:
\begin{equation}
P_{\text{fuse}} = P^{(l-1)} + \mathrm{Linear}\big( [ \tanh(g) \odot P_{\text{raw}} ; P_{\text{ego}} ; P_{\text{sa}} ] \big).
\end{equation}

\subsubsection{Visual Spatial Refinement}
While $P_{\text{fuse}}$ successfully captures high-level semantic intent, directly decoding it often yields physically infeasible trajectories due to the inherent lack of explicit spatial boundaries in language models. To rectify this, the second stage introduces a rigorous spatial refinement mechanism. By utilizing the dense geometric features $V_{\text{spatial}}$ extracted from the Spatial Stream, we inject explicit spatial and geometric priors into the network. This forces the planning tokens to strictly adhere to real-world motion constraints and obstacle boundaries:
\begin{equation}
\begin{aligned}
P_{\text{vis}} &= P_{\text{fuse}} + \mathrm{MHCA}_3(P_{\text{fuse}}, V_{\text{spatial}}, V_{\text{spatial}}), \\
P^{(l)} &= P_{\text{vis}} + \mathrm{FFN}(P_{\text{vis}}).
\end{aligned}
\end{equation}

\subsubsection{Trajectory Decoding and Multi-Task Objective}
After $N$ cascaded blocks, the final planning tokens $P^{(N)}$ are processed by a lightweight MLP decoder to predict the future ego-vehicle trajectory $\hat{Y} \in \mathbb{R}^{M \times 3}$. The trajectory prediction is supervised against the ground-truth trajectory $Y_{\text{gt}}$ using an $L_1$ regression loss. 
To ensure physical feasibility and rider comfort, this objective is augmented with kinematic smoothness constraints that explicitly penalize excessive acceleration ($a$) and jerk ($j$):
\begin{equation}
\mathcal{L}_{\text{plan}} = \mathcal{L}_{L1}(\hat{Y}, Y_{\text{gt}}) + \lambda_{\text{smooth}} \sum_{x \in \{a, j\}} \mathcal{L}_{\text{SmoothL1}}(x, 0).
\end{equation}

The entire $\mathrm{S}^2$-VLA framework is trained end-to-end using a joint multi-task objective function that balances planning, dynamic agent perception, and static map grounding:
\begin{equation}
\mathcal{L}_{\text{total}} = \lambda_{\text{plan}}\mathcal{L}_{\text{plan}} + \lambda_{\text{agent}}\mathcal{L}_{\text{agent}} + \lambda_{\text{map}}\mathcal{L}_{\text{map}}.
\end{equation}

\section{Experiments}

\subsection{Dataset and Metrics}
We conduct end-to-end planning training and evaluation on the NAVSIM benchmark~\cite{dauner2024navsim}, which emphasizes closed-loop interactions between the agent and dynamic environments.

NAVSIM introduces the Predictive Driver Model Score (PDMS) as a comprehensive metric for evaluating closed-loop performance, showing strong correlation with standard driving metrics. PDMS is computed based on five criteria: No Collision (NC), Drivable Area Compliance (DAC), Time to Collision (TTC), Comfort (Comf), and Ego Progress (EP).

\vspace{-1em}
\subsection{Implementation Details}

We employ a three-stage training procedure. First, we perform Supervised Fine-Tuning (SFT) on the pre-trained VLM using the ReCogDrive VQA dataset~\cite{li2025recogdrive} for 3 epochs, significantly enhancing its instruction-following and spatial reasoning capabilities in complex traffic scenarios. Second, with the VLM backbone frozen, we apply LoRA~\cite{hu2022lora} and train the intent extraction and auxiliary perception modules from scratch for 4 epochs, temporarily decoupling the visual spatial features from the Dual-Stream Planning Adapter. In the final stage, we freeze the perception and intent components, dedicating 4 epochs of optimization exclusively to the visual refinement module to align generated trajectories with precise spatial constraints.
The entire framework is optimized for end-to-end trajectory planning on the NAVSIM benchmark~\cite{dauner2024navsim}. All experiments are conducted using the AdamW optimizer on 4 NVIDIA A100 GPUs. Detailed hyperparameters are summarized in Table~\ref{tab:hyperparameters}.

\vspace{-1em}
\begin{table}[htbp]
    \centering
    \caption{HYPERPARAMETER SETTINGS FOR $S^2$-VLA.}
    \label{tab:hyperparameters}
    \begin{tabular}{ll}
        \toprule
        \textbf{Hyperparameter} & \textbf{Value} \\
        \midrule
        \multicolumn{2}{l}{\textbf{Training Setup}} \\
        Learning Rate & $1 \times 10^{-4}$ \\
        Batch Size & 16 \\
        LoRA Rank ($r$) / Alpha ($\alpha$) & 8 / 16 \\
        \midrule
        \multicolumn{2}{l}{\textbf{Loss Weights}} \\
        $\lambda_{\text{plan}}$ & 1.0 \\
        $\lambda_{\text{agent}}$ & 0.1 \\
        $\lambda_{\text{map}}$ & 0.5 \\
        $\lambda_{\text{smooth}}$ & 0.5 \\
        $\lambda_{\text{reg}}$ & 1.0 \\
        $\lambda_{\text{cls}}$ & 1.0 \\
        $\omega_{c}$ & 1.0 \\
        \bottomrule
    \end{tabular}
\end{table}

\section{Results}

\subsection{Quantitative Results}
We evaluate $\mathrm{S}^2$-VLA against state-of-the-art E2E, VLMs and  VLA methods on the NAVSIM closed-loop planning benchmark. 
To ensure a rigorous and fair comparison, all models reported in Table~\ref{tab:navsim} are evaluated strictly under their respective supervised training paradigms (i.e., behavioral cloning for E2E models and supervised fine-tuning for VLMs/VLAs). 
We intentionally exclude post-training strategies from the baselines, such as closed-loop reinforcement learning via Group Relative Policy Optimization (GRPO), to ensure a fair comparison of the network designs.

\begin{table*}[!tbp]
    \centering
    \caption{Performance comparison on the NAVSIM navtest benchmark.}
    \label{tab:navsim}
    \begin{threeparttable} 
        \begin{tabular}{lccc|ccccc>{\columncolor{gray!20}}c} 
            \toprule
            \textbf{Method} & \textbf{Publication} & \textbf{Image} & \textbf{Lidar} & \textbf{NC$\uparrow$} & \textbf{DAC$\uparrow$} & \textbf{EP$\uparrow$} & \textbf{TTC$\uparrow$} & \textbf{Comf$\uparrow$} & \textbf{PDMS$\uparrow$} \\
            \midrule
            \textbf{E2E} & & & & & & & & & \\
            TransFuser \cite{chitta2022transfuser}  & IEEE T. PAMI, 2023 & $\surd$ & $\surd$ & 97.8 & 92.6 & 78.9 & 92.9 & \textbf{100} & 83.9 \\
            UniAD \cite{hu2023uniad} & CVPR, 2023 & $\surd$ &  & 97.8 & 91.9 & 78.8 & 92.9 & \textbf{100} & 83.4 \\
            PARA-Drive \cite{weng2024para-drive} & CVPR, 2024 & $\surd$ &  & 97.9 & 92.4 & 79.3 & 93.0 & 99.8 & 84.0 \\
            DRAMA \cite{yuan2024drama} & arXiv, 2024 & $\surd$ & $\surd$ & 98.2 & 95.2 & 81.3 & 94.2 & \textbf{100} & 86.9 \\
            ARTEMIS \cite{feng2025artemis} & IEEE RA-L, 2025 & $\surd$ & $\surd$ & 98.3 & 95.1 & 81.4 & 94.3 & \textbf{100} & 87.0 \\
            DiffusionDrive \cite{liao2025diffusiondrive} & CVPR, 2025 & $\surd$ & $\surd$ & 98.2 & \textbf{96.2} & \textbf{82.2} & \textbf{94.7} & \textbf{100} & \textbf{88.1} \\
            \midrule
            \textbf{VLM/VLA} & & & & & & & & & \\
            InternVL3-2B \cite{zhu2025internvl3} & arXiv, 2025 & $\surd$ &  & 97.6 & 93.1 & 79.1 & 92.7 & \textbf{100} & 84.1 \\
            ImagiDrive \cite{li2025imagidrive} & ICRA, 2026(to appear) & $\surd$ &  & 97.9 & 95.5 & 80.7 & 93.1 & 99.9 & 86.4 \\
            AutoVLA\tnote{*}  \cite{zhou2025autovla} & NeurIPS, 2025 & $\surd$ &  & 96.9 & 94.4 & 75.8 & 88.1 & 99.9 & 80.5 \\
            ReCogDrive\tnote{*}  \cite{li2025recogdrive}  & ICLR, 2026(to appear) & $\surd$ &  & 98.1 & 94.7 & 80.9 & 94.2 & \textbf{100} & 86.5 \\
            $\mathrm{S}^2$-VLA(\textbf{Ours}) & - & $\surd$ &  & \textbf{98.4} & 94.9 & 81.6 & 94.6 & \textbf{100} & 87.1 \\
            \bottomrule
        \end{tabular}
        
        \begin{tablenotes}
            \footnotesize
            \item[*] The RL-augmented variants of these models are excluded to ensure a fair, SFT-only comparison. 
        \end{tablenotes}
    \end{threeparttable}
\end{table*}
\vspace{-1em}
\begin{table*}[!hbtp]
    \centering
    \caption{Ablation study of key components.}
    \label{tab:ablation_study}
    \begin{tabular}{cccc|ccccc>{\columncolor{gray!20}}c}
        \toprule
         \textbf{BaseVLM} & \textbf{Semantic Feature} & \textbf{Spatial Feature} & \textbf{Aux. Perception} & \textbf{NC}$\uparrow$ & \textbf{DAC}$\uparrow$ & \textbf{EP}$\uparrow$ & \textbf{TTC}$\uparrow$ & \textbf{Comf}$\uparrow$ & \textbf{PDMS}$\uparrow$ \\
        \midrule
         $\surd$ & & & & 97.6 & 93.1 & 79.1 & 92.7 & 100 & 84.1 \\
         $\surd$ & $\surd$ & & & 98.2 & 94.0 & 80.2 & 93.9 & 100 & 85.6 \\
         $\surd$ & $\surd$ & $\surd$ & & 98.1 & 94.5 & 80.6 & 94.2 & 100 & 86.2 \\
         $\surd$ & $\surd$ & $\surd$ & $\surd$ & 98.4 & 94.9 & 81.6 & 94.6 & 100 & 87.1 \\
        \bottomrule
    \end{tabular}
    
\end{table*}

\vspace{1em}

\textbf{Comparison with E2E Models.}
As shown in Table~\ref{tab:navsim}, $\mathrm{S}^2$-VLA achieves a PDMS of 87.1 using only a single front-view camera. It substantially outperforms vision-only counterparts such as UniAD (83.4) and PARA-Drive (84.0), and even surpasses established LiDAR-fused methods like ARTEMIS (87.0) and DRAMA (86.9). This  indicates that our dedicated spatial stream effectively infers robust 3D structural constraints directly from 2D imagery, significantly mitigating the inherent depth ambiguity of monocular vision.

While DiffusionDrive achieves a marginally higher PDMS of 88.1, maintaining a slight lead over our method primarily in Drivable Area Compliance, it is important to note the architectural differences. DiffusionDrive relies on an iterative, computationally heavy diffusion decoder and explicit LiDAR point clouds, which naturally provide absolute physical boundaries to circumvent 2D scale ambiguity. The fact that $\mathrm{S}^2$-VLA achieves highly competitive performance without relying on expensive 3D sensors or iterative decoding underscores the sheer representational efficiency of our decoupled semantic-spatial architecture.

\textbf{Comparison with VLM/VLA Models.}
Compared to a naive VLM baseline, which utilizes the same InternVL3-2B backbone but relies on autoregressive text-head decoding for trajectory prediction,  $\mathrm{S}^2$-VLA improves PDMS by 3.0 points. This substantial gain directly exposes the semantic-geometric misalignment problem inherent in standard VLMs: when trajectories are generated solely via discrete language tokens, continuous spatial information is inevitably quantized and degraded, leading to geometrically imprecise plans. Our dual-stream design resolves this by routing geometric reasoning through a dedicated spatial pathway, successfully preserving the structural fidelity that language abstraction otherwise discards.
Furthermore, $\mathrm{S}^2$-VLA outperforms recent methods like ReCogDrive (86.5) and ImagiDrive (86.4), which attempt to enhance VLA planning through enriched textual reasoning chains or imagination-augmented training. Notably, our model achieves a higher PDMS despite utilizing a significantly smaller backbone.

A breakdown of the sub-metrics further highlights the nature of these improvements. $\mathrm{S}^2$-VLA achieves the highest No Collision and Ego Progress scores among all evaluated VLM/VLA  methods, reflecting superior collision avoidance and forward progress. These safety-critical gains stem directly from explicit spatial grounding: the injected geometric priors enable highly accurate perception of obstacle proximity and road curvature, which are critical physical boundaries where discrete token representations fundamentally fail.

\subsection{Qualitative Analysis}

\begin{figure*}[!htbp]
    \centering
        \vspace{-0.5cm}
     \includegraphics[width=1\textwidth]{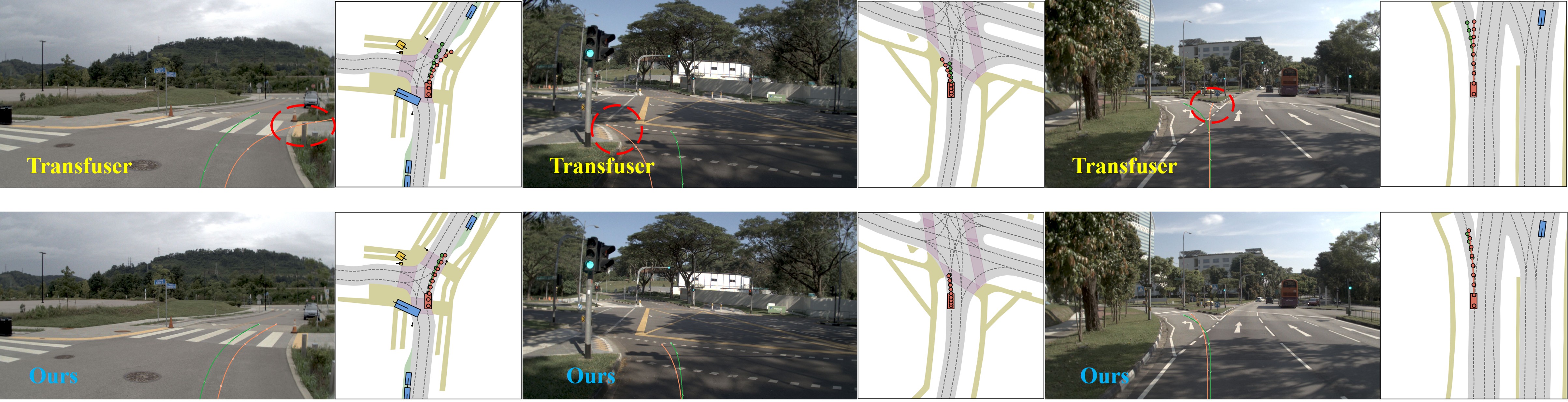}
    (a)  Qualitative comparison with Transfuser.

    \vspace{0.4cm}
    \includegraphics[width=1\textwidth]{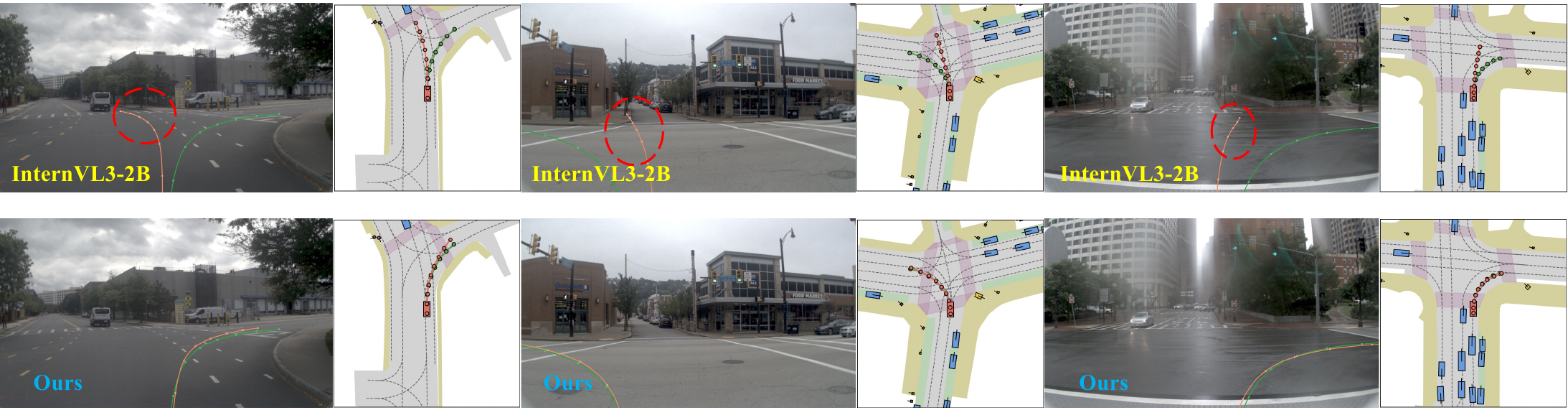}
    (b) Qualitative comparison with InternVL3-2B.

    \vspace{0.4cm}
    \includegraphics[width=1\textwidth]{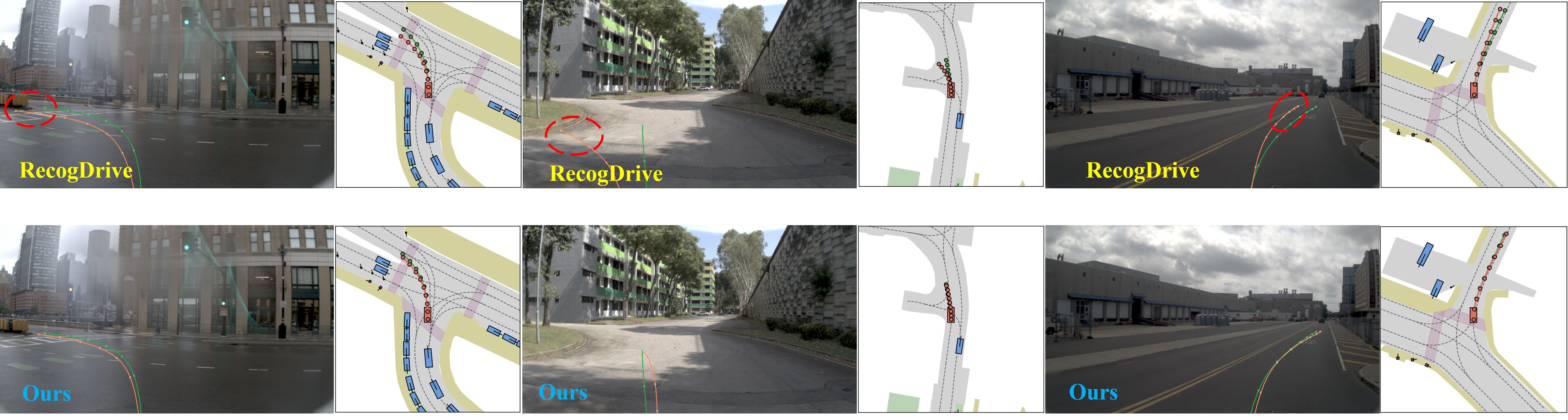}
    (c) Qualitative comparison with RecogDrive.
       \caption{Qualitative analysis. The green curves represent ground-truth expert trajectories, and the orange curves represent the model's predicted future trajectories.
}
    
    \vspace{-0.3cm}
    \label{fig:qualitative_analysis}
\end{figure*}

We present qualitative trajectory comparisons between $\mathrm{S}^2$-VLA and three representative baselines on the NAVSIM navtest benchmark in Fig.~\ref{fig:qualitative_analysis}.

Despite having access to explicit LiDAR data, the E2E Transfuser\cite{chitta2022transfuser} baseline exhibits unsafe behaviors, such as curb encroachment and boundary violations in complex road topologies (Fig.~\ref{fig:qualitative_analysis}(a)). This indicates that raw multi-modal sensor fusion, without sufficient high-level semantic understanding, struggles in scenarios that demand a synthesis of geometric precision and contextual reasoning.

InternVL3-2B \cite{zhu2025internvl3}, the VLM  baseline exhibits systematic geometric deviations during high-curvature turns and at intersections (red circles in Fig.~\ref{fig:qualitative_analysis}(b)), frequently failing to maintain lane centering. This failure pattern highlights the inherent limitation of representing continuous spatial geometry via discrete language tokens: the quantization of spatial coordinates into vocabulary entries introduces irrecoverable rounding errors. These errors inevitably accumulate along the trajectory horizon, producing progressively larger deviations from the intended path.

ReCogDrive\cite{li2025recogdrive}  attempts to improve planning by injecting VLM-derived cognitive tokens into a diffusion-based planner. However, its long-range predictions still diverge from the ground truth, particularly in high-curvature turns and intersections lacking clear lane markings (Fig.~\ref{fig:qualitative_analysis}(c)). This perfectly illustrates our core motivation: the deep feature compression inherent to standard VLM backbones inherently destroys the fine-grained spatial details and structural priors required for accurate, boundary-aware trajectory planning.

In contrast, $\mathrm{S}^2$-VLA consistently generates trajectories that closely align with expert demonstrations across diverse and complex scenarios. The planned paths remain smooth, stable, and strictly adherent to physical road structures, even over long-range horizons featuring occlusions or degraded lane markings. This robustness directly validates our core architectural hypothesis: by explicitly decoupling high-level semantic reasoning from low-level geometric perception, and seamlessly fusing them via the Dual-Stream Planning Adapter, the model successfully marries contextual scene understanding with precise spatial grounding. Ultimately, $\mathrm{S}^2$-VLA achieves a synergistic balance of capabilities that neither a pure VLM nor a pure geometric planner can provide alone.

\vspace{-1 em}

\subsection{Ablation Study}

Table~\ref{tab:ablation_study} presents a systematic ablation study that isolates the contribution of each component, highlighting their contribution in semantic reasoning, geometric precision, and spatial grounding, and thereby validating the design of $\mathrm{S}^2$-VLA.

Introducing the planning adapter with multi-level semantic features improves PDMS by +1.5 (84.1 $\rightarrow$ 85.6). The gains are most pronounced in Ego Progress (EP, +1.1) and Time to Collision (TTC, +1.2), indicating enhanced long-horizon planning and hazard anticipation. 
By capturing both high-level intent and contextual scene understanding, the hierarchical semantic features enable better coordination between immediate actions and long-term objectives. However, the relatively modest gain in Drivable Area Compliance (DAC, +0.9) confirms that semantic reasoning alone is insufficient for precise geometric alignment.

To address this, adding the ViT-based spatial stream further increases the PDMS to 86.2 (+0.6), with the most significant improvement observed in DAC (94.0 $\rightarrow$ 94.5). This highlights the complementary role of the spatial stream: by preserving fine-grained geometric details via spatially dense visual features, it explicitly enforces structural alignment. Consequently, the model achieves superior lane adherence and boundary awareness, effectively mitigating the geometric limitations of purely semantic representations. Achieving this empirical gain as a residual refinement atop a converged semantic baseline also validates our cascaded architectural ordering. 
  The design mirrors the natural factorization of trajectory planning, which establishes intent before verifying geometric feasibility. A parallel alternative would instead force planning tokens to compete across heterogeneous keys, diluting each stream's specialization and recreating the very entanglement $\mathrm{S}^2$-VLA is designed to prevent.

Finally, incorporating auxiliary perception tasks yields a further +0.9 PDMS gain, reaching a final score of 87.1. Notably, the largest improvements occur in safety-critical metrics, including No Collision (NC, +0.3) and EP (+1.0). These gains are driven by explicit multi-task supervision comprising both BEV map generation and agent prediction. 
 This dual objective ensures that the latent representations successfully capture both static geometric boundaries and dynamic obstacle trajectories, providing a holistic physical prior for the downstream planning adapter.

\section{Conclusion}
This letter introduced $\mathrm{S}^2$-VLA, a Semantic-Spatial Dual-Stream Vision-Language-Action framework designed for end-to-end autonomous driving. To bridge the fundamental gap between high-level semantic reasoning and low-level physical execution in standard VLMs, we developed a dedicated multi-modal planning adapter. This adapter seamlessly fuses multi-scale semantic features from the language backbone with uncompressed, fine-grained geometric features extracted directly from the visual encoder. By further integrating BEV agent state prediction and semantic mapping as auxiliary supervision, we explicitly inject spatial and geometric  priors into the network. Evaluations on the NAVSIM benchmark demonstrate that $\mathrm{S}^2$-VLA successfully mitigates the spatial perception deficiencies of traditional VLMs, achieving highly competitive closed-loop performance under pure supervised fine-tuning. Ultimately, this research confirms that preserving continuous spatial features, bypassing the quantization bottleneck of discrete language heads, is critical for generating safe and dynamically feasible driving trajectories.

Despite its effectiveness, our framework has two limitations, each suggesting a natural direction for future work. First, the dense extraction and dual-stream fusion of spatial features incur non-trivial computational overhead; we aim to develop sparser feature-representation mechanisms that replace global attention fusion and inject richer geometric priors at lower cost. Second, because our dual-stream architecture is orthogonal to the training paradigm, augmenting it with closed-loop reinforcement learning post-training offers a natural and compatible next step.

\bibliography{ral2025_vla}

\end{document}